# Spread Unary Coding

Subhash Kak[1]

**Abstract**
Unary coding is useful but it is redundant in its standard form. Unary coding can also be seen as spatial coding where the value of the number is determined by its place in an array. Motivated by biological finding that several neurons in the vicinity represent the same number, we propose a variant of unary numeration in its spatial form, where each number is represented by several 1s. We call this spread unary coding where the number of 1s used is the spread of the code. Spread unary coding is associated with saturation of the Hamming distance between code words.

## 1. Introduction

The usefulness of unary coding in certain situations arises out of the fact that the Hamming distance between numbers increases linearly with respect to the difference between them. The unary code for the number *n* is a sequence of *n* 1s. To mark the beginning of a new unary number one might append a 0 to its left. Since the length of the codes increases with each number, such codes are optimal if the probabilities of the numbers are correspondingly decreasing in powers of 2 and they have been used in data compression [1],[2].

The distance property of the normal form of the unary code is that the Hamming distance, $d_a$, between two numbers $n_1$ and $n_2$ equals the difference between the numbers:

$$d_a(n_1 - n_2) = n_1 - n_2 \qquad (1)$$

In contrast the binary representation of numbers has no similarly uniform distance property. For example, the Hamming distance between 0 and any power of 2 in a binary code is only 1. In the Gray code, successive numbers have a distance of 1, but the property does not extend beyond numbers in succession. Due to such nonlinear behavior, binary coding does not possess inherent error correction. In this binary coding is different from unary coding for which inherent error correction capacity was recently investigated [3].

The linearity of the Hamming distance makes it easier to separate patterns in instantaneously trained neural network training [4]-[10]. But this coding, while effective in computer learning, does not take into account the actual manner in which such learning is done in biological neural networks. Furthermore, we have no evidence that the linearity of the distance measure beyond a certain maximum provides any specific advantage as far as effectiveness of a learning algorithm is concerned.

The role of unary coding in biology has become clear in the study of avian birdsong production [11]-[17]. The HVC (high vocal center) is a nucleus in the brain of the songbirds that plays a part in both the learning and the production of bird song. It is located in the lateral caudal nidopallium and has projections to both the direct and the anterior forebrain pathways [11]-[18]. A site of characteristic pre-motor activity in songbirds is the forebrain robust nucleus of the archistriatum (RA) that is associated with the generation of stereotyped sequences of spike bursts during song and recapitulation of these sequences during sleep.

---
[1] Oklahoma State University, Stillwater, OK 74078

According to Hahnloser et al. [12], "recordings of identified HVC neurons in sleeping and singing birds show that individual HVC neurons projecting onto RA neurons produce bursts sparsely, at a single, precise time during the RA sequence. These HVC neurons burst sequentially with respect to one another." The authors suggest "that at each time in the RA sequence, the ensemble of active RA neurons is driven by a subpopulation of RA-projecting HVC neurons that is active only at that time. As a population, these HVC neurons may form an explicit representation of time in the sequence." The fact of unary numeration is a consequence of this spatial coding.

Several neurons in the vicinity represent the same number which may code the relative time at which a specific song sub-sequence begins. This also indicates that the unary number is not represented by a single neuron, but rather a collection of neurons.

This paper considers the mathematical basis of a multi-neuron unary code which will be called the *spread unary* code.

2. **Normal Unary Codes**

Table 1 shows the standard form of unary code in first column and its representation as a spatial code in second column.

Table 1. Unary code in normal and spatial forms

| n | Normal form | Spatial form |
|---|---|---|
| 0 | 0 | 0000000000 |
| 1 | 01 | 0000000001 |
| 2 | 011 | 0000000010 |
| 3 | 0111 | 0000000100 |
| 4 | 01111 | 0000001000 |
| 5 | 011111 | 0000010000 |
| 6 | 0111111 | 0000100000 |
| 7 | 01111111 | 0001000000 |
| 8 | 011111111 | 0010000000 |
| 9 | 0111111111 | 0100000000 |
| 10 | 01111111111 | 1000000000 |

The normal form has redundancy which is why it can be used for certain inherent error correction, but this redundancy can be costly, which is why this form is not found in Nature. The spatial form is efficient (if 0s are replaced by "no signal") but it is not robust since its distance ($d_b$) property is:

$$d_b(n_1 - n_2) = 2 \tag{2}$$

The lack of robustness is the reason it is not suitable for practical applications.

We wish to develop a robust variant of the spatial unary code that guarantees a large distance between the codes. We propose a couple of variants in this paper.

3. **Spread Unary Codes**

*First Code:* The first spread unary code will replace each 1 in the spatial form with *k* 1s. Thus the number *n* will be represented by *k* 1s followed by *n-1* 0s:

*n: 11... (*k times*)00... (*n-1 times*)* (3)

Let the distance between the code words for this unary code be represented by $d_{s1}$. Since each number has $k$ 1s, numbers that are next to each other will have distance of 2 and thereafter the distance will increase in steps of 2 until the difference between the numbers is $k$ or higher when the distance will remain fixed at $2k$. Unlike the normal unary case given by (1), we have saturation of distance that is shown in Figure 1.

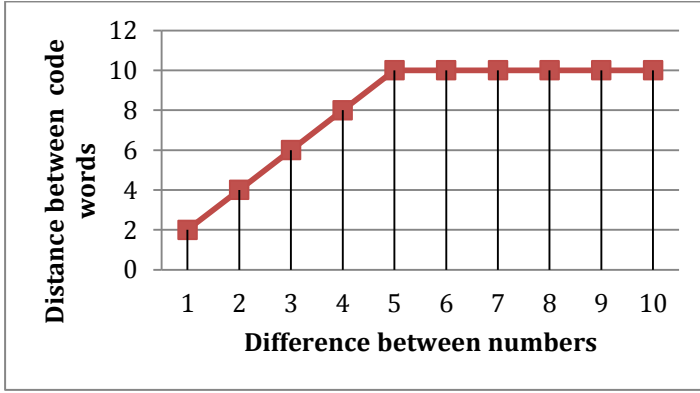

Figure 1. The distance between code words for k=5

**Theorem.** The distance between the code words according to (3) of two non-zero numbers $n_1$ and $n_2$ is given by the following expression:

$$d_{s1} = \begin{cases} 2(n_2 - n_1), & \text{when } n_2 - n_1 < k \\ 2k, & \text{when } n_2 - n_1 \geq k \end{cases} \quad (4)$$

*Example 1.* Table 2 presents an example of the representation of numbers 0 through 10 for k=3.

Table 2. First spread unary code for k=3 for numbers 1 through 10

| n | Spread unary code |
|---|---|
| 0 | 000000000000 |
| 1 | 000000000111 |
| 2 | 000000001110 |
| 3 | 000000011100 |
| 4 | 000000111000 |
| 5 | 000001110000 |
| 6 | 000011100000 |
| 7 | 000111000000 |
| 8 | 001110000000 |
| 9 | 011100000000 |
| 10 | 111000000000 |

Note that the value of the number in Table 2 is according to the location of the rightmost 1. But in a variant case, one may choose the location of the number as the midpoint of the group of 1s, where for simplicity it is assumed that $k$ is odd. This requires that the count for location be offset to the left by $\lfloor (k-1)/2 \rfloor$. In the case of the above example, that is an offset by 1 to the left.

The minimum and the maximum distance range between 2 and 6 for this example.

*Second Code:* The second spread unary code will replace each 1 in the spatial form with *k* numbers of some chosen property. For example, one could use numbers that increase and then decrease with the peak at the middle, the putative label of the number:

*n: 123…321 (*k digits*)00… (*n-1 digits*)* (5)

Let the distance between the code words for this unary code be $d_{s2}$, where the distance is defined as sum of the absolute value of the difference of the digits. For odd *k*, the distance between adjacent nonzero numbers will be *k*+1; for two numbers that have a difference of 2, the distance will be 2*k*; for a difference of 3, the distance is 3*k*-1, and so on. When the separation between the numbers equals *k* or greater, the distance between the code words is $(k+1)^2/2$.

*Example 2.* Table 3 presents an example of the representation of numbers 0 through 10 for k=5 for the second type of spread unary code.

Table 3. Second spread unary code for k=5 for numbers 1 through 9

| n | Spread unary code |
|---|---|
| 0 | 0000000000000 |
| 1 | 0000000012321 |
| 2 | 0000000123210 |
| 3 | 0000001232100 |
| 4 | 0000012321000 |
| 5 | 0000123210000 |
| 6 | 0001232100000 |
| 7 | 0012321000000 |
| 8 | 0123210000000 |
| 9 | 1232100000000 |

The maximum distance between any two code words is 18. Clearly with its larger distance between code words, the second type will be more resistant to noise and errors.

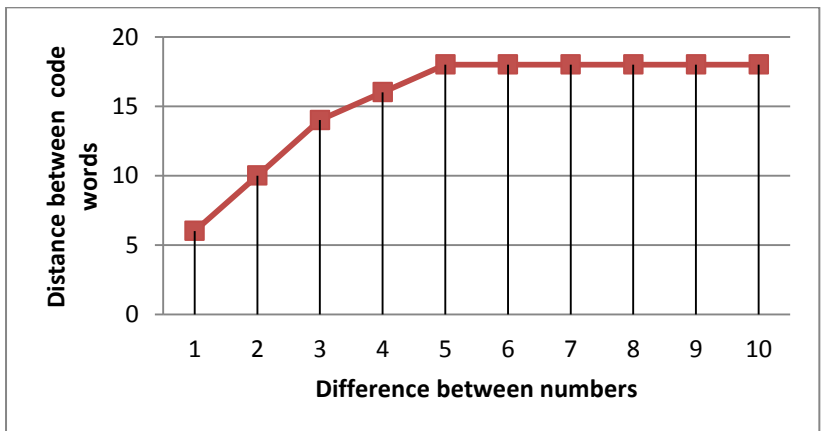

Figure 2. The distance between code words for k=5 for second type

In the case of biological coding, the neuron with the highest output will be the putative neuron corresponding to the number that is defined as the central point of the area.

## Conclusions

Motivated by biological finding that several neurons in the vicinity represent the same number, we propose a variant of unary numeration in its spatial form, where each number is represented by several 1s. We call this spread unary coding where the number of 1s used is the spread of the code.

Spread unary coding is associated with the property of distance saturation beyond the spread. Since some form of spread unary coding appears to be in use in biological systems, it is likely that such coding will also be useful in artificial neural networks and other applications.